\documentclass[14pt]{article}

\usepackage{natbib}

\usepackage{url}

\usepackage{graphicx} % more modern

\usepackage[version=3]{mhchem}

\usepackage{amssymb,amsmath,bm}

\usepackage{lscape}

\usepackage{color}

\usepackage{datetime}

\renewcommand{\vec}[1]{\bm{#1}}

\title{Prediction of peptide bonding affinity: kernel methods for nonlinear modeling}
\author{Charles Bergeron\thanks{E-mail address: chbergeron@gmail.com} \\
                Department of Mathematical Sciences\\
        \and
        Theresa Hepburn, C. Matthew Sundling, Michael Krein \\
        Bill Katt, Nagamani Sukumar,Curt M. Breneman \\
                Center for Biotechnology and Interdisciplinary Studies \\
        \and
        Kristin P. Bennett \\
                Departments of Mathematical Sciences and Computer Science\\
                \and
                Rensselaer Polytechnic Institute\\
                Troy, NY, 12180
}
\date{\today}

\begin{document}
\maketitle

\begin{abstract}

This paper presents regression models obtained from a process of blind prediction of peptide binding affinity from provided descriptors for several distinct datasets as part of the 2006 Comparative Evaluation of Prediction Algorithms (COEPRA) contest. This paper finds that kernel partial least squares, a nonlinear partial least squares (PLS) algorithm, outperforms PLS, and that the incorporation of transferable atom equivalent features improves predictive capability. 
\\
List of acronyms: Comparative Evaluation of Prediction Algorithms (COEPRA), cross-validation (CV), kernel partial least squares (KPLS), leave-one-out (LOO), partial least squares (PLS), reproducing kernel Hilbert space (RKHS), Transferable Atom Equivalent (TAE), two-dimensional (2D)

\end{abstract}

\section{Introduction}

Comparative Evaluation of Prediction Algorithms (COEPRA, \url{http://www.coepra.org/}) is a modeling competition organized to provide objective testing of various algorithms via the process of blind prediction for chemical, biological, and medical data. COEPRA's stated goals are to advance modeling algorithms and software as well as provide reference datasets to the research community.

Transferable Atom Equivalent (TAE) RECON features are electron-density derived descriptors obtained by fragment reconstruction. MOE features are geometrical, structural, physiochemical and topological 2D descriptors. RAD features are topological autocorrelations of TAE RECON descriptors. This paper shows how their addition to the COEPRA descriptors improve modeling performances.

Partial least squares (PLS) regression is a machine learning technique. Because it considers the covariance of the inputs to the output to guide the selection of features, it is much more stable than multiple linear regression. This approach was developed for, and is popular with, the field of chemometrics where the number of variables is much greater than the number of samples, and where a high occurrence of correlated representations exists [1]. Less well-known to the chemometrics community is kernel partial least squares (KPLS) regression, a generalization of PLS that finds a nonlinear relation between features instead of being limited to a linear combination thereof [2]. This paper demonstrates how KPLS largely outperforms PLS in the COEPRA datasets.

The remainder of this paper accomplishes the following:
\begin{itemize}
  \item The COEPRA datasets are described.
  \item TAE RECON features are defined, and SIMIL scores are introduced.
  \item PLS is generalized to a nonlinear, KPLS framework.
  \item Implementation issues for KPLS are discussed.
  \item Models submitted to the contest and their performances are stated.
  \item Post-contest analysis of the datasets resulting in new, better performing models is presented.
  \item Conclusions for the paper are provided.
\end{itemize}

\section{Notation}

Let $\vec{x}$ denote a column vector. Let $\vec{x}^T$ denote the transpose of $\vec{x}$. Let $X$ denote a matrix, $X^T$ it's transpose and $X^{-1}$ it's inverse. Write the identity matrix of appropriate size as $I$. The expected value of a set of number assembled into vector $\vec{x}$ is written $E(\vec{x})$ and represents the mean value: $E(\vec{x})=\frac{1}{n}\sum_{i=1}^{n}{x_i}$. The Euclidean norm $\|\vec{x}\|_2$ of $\vec{x}$ is given by $\|\vec{x}\|_2=\sqrt{\sum_{i=1}^{n}{x_i^2}}$.

\section{COEPRA datasets}

Three regression tasks were proposed during the summer of 2006. For each task, a dataset consists of a calibration set and a prediction set. The following data are provided:
\begin{itemize}
  \item For the calibration set, the COEPRA descriptors and the corresponding responses for each sample.
  \item For the prediction set, the COEPRA descriptors for each sample.
\end{itemize}
Hence, the calibration set is used to develop a model, and this model is propagated to prediction set descriptors to make a prediction for the response. These predictions are compared with the actual values by the contest organizers after the close of the submission date.

Each sample consists of a peptide sequence of amino acid residues (rounds 1 and 3 involve nonapeptides while round 2 involves octapeptides) and 643 COEPRA descriptors per amino acid. The nature of these descriptors remain unknown to this time. Table \ref{table1} presents basic characteristics of each round.

\begin{table}[tb!]									
\caption{Basic information about the datasets.}\label{table1}									
\begin{center}									
\begin{tabular}{ccccc}									
\hline									
\bf{dataset}	&	\bf{calibration}	&	\bf{prediction}	&	\bf{amino}	&	\bf{COEPRA}	\\
	&	\bf{samples}	&	\bf{prediction}	&	\bf{acids}	&	\bf{descriptors}	\\
\hline									
1	&	89	&	88	&	9	&	5787	\\
2	&	76	&	76	&	8	&	5144	\\
3	&	133	&	133	&	9	&	5787	\\
\hline									
\end{tabular}									
\end{center}									
\end{table}															

During the contest, the nature of the regression value was not known. That is no longer the case. For round 1, the output is the bonding affinity to HLA-A*0201 major histocompability complex [3]. For round 2, the output is the binding affinity of mouse class I major histocompability complex [4]. For round 3, the output is the bonding affinity to HLA-A*0201 major histocompability complex [5].

The contest methodology proposed, for each round, to learn a model from calibration data, for which the response is known, and then propagate that model to prediction data to establish a prediction of the response that could then be evaluated by contest organizers against the true values. Contestants were free to add additional descriptors; thus we augmented the COEPRA descriptors with MOE and RAD descriptors that we now describe.

\section{Atomic charge density fragment features}

RECON is an algorithm for the rapid reconstruction of molecular electron densities and electron density-based properties of molecules, using pre-computed atomic charge density fragments and associated descriptors stored in a Transferable Atom Equivalent library. Molecular TAE descriptors are constructed in most cases by summation of the respective atomic fragment contributions. The TAE technology [6,7,8,9] provides a rapid means of computing electronic property information for large molecular datasets. Among the descriptors used in this study are traditional 2D MOE descriptors and topological RECON autocorrelation descriptors (RAD), which are autocorrelations of surface integrals of various electron density derived (TAE) atomic properties $P_x,P_y$:
\begin{equation}
A(R_{xy})=\frac{1}{n}\sum_{(x=1}^{n}{\sum_{(y=1}^{n}{P_x P_y}}
\end{equation}
binned by the minimum bond path $R_{xy}$ (topological distance) between the respective pair of atoms $(x,y)$. Use of the minimum bond path allows topological autocorrelation descriptors to be computed without the need for three-dimensional energy minimized structures [10]. The electron density-derived properties used are the electrostatic potential, the electronic kinetic energy density, gradients of the electron density and electronic kinetic energy density normal to an electron density isosurface (corresponding to the molecular van der Waals surface), the Fukui function, the Laplacian distribution of the electron density, the bare nuclear potential and a local average of the ionization potential on the surface. These features have been described in detail elsewhere [6,7,8,9] (online version for proteins and polypeptides available at \url{http://reccr.chem.rpi.edu/}). The implementation [10] of the RECON algorithm within MOE is used in this study.

SIMIL scores are a new type of similarity score, new to this study, between pairs of amino acid residues. Each SIMIL score is a two-part score, consisting of a Class Score and a RECON Score. The Class Score is a weighted score constructed out of bits representing the presence of the following physical characteristics: tiny, small, positive, negative, polar, non-polar, aliphatic, and aromatic. The RECON Score is constructed from weighted differences of TAE RECON descriptors. These SIMIL scores appear as a 20-by-20 similarity matrix.

\section{Machine learning methods}

Linear regression is based on the notion of the dot-product functions in the Euclidean space. For example, each entry of the covariance matrix $C$ is given by the dot-product function:
\begin{equation}
c(\vec{x},\bar{\vec{x}})=\vec{x}^T\bar{\vec{x}}.
\end{equation}
Nonlinear regressions can be achieved by using other functions, called kernel functions, that satisfy dot-product properties in a different space called a reproducing kernel Hilbert space (RKHS). A linear regression can be computed in RKHS that is usually of much higher (and possibly infinite) dimensionality, resulting in a model that is nonlinear in Euclidean space. Each entry of the kernel matrix $K$, of same size as $C$, requires one evaluation of the kernel function. The fact that the computational effort of working in a potentially infinite-dimensional space is capped by the number of samples is called the kernel trick.

The combination of PLS with kernels produces a powerful algorithm: kernel partial least squares regression [2]. The model is of the form
\begin{equation}
K\vec{\beta} \approx \vec{y}
\end{equation}
where $K$ is a square kernel matrix whose size is the number of samples computed from the features and $\vec{y}$ is a vector of responses. The most-oft cited kernel function is the Gaussian one, given by
\begin{equation}
k(\vec{x},\bar{\vec{x}})=\exp{\left(-\frac{\|\vec{x}-\bar{\vec{x}}\|_2^2}{2\eta^2}\right)}
\end{equation}
where $\vec{x},\bar{\vec{x}}$ are sample vectors. A variant is the exponential kernel:
\begin{equation}
k(\vec{x},\bar{\vec{x}})=\exp{\left(-\frac{\|\vec{x}-\bar{\vec{x}}\|_2}{2\eta}\right)}.
\end{equation}
Working with either kernel requires setting parameter $\eta$.

The vector of coefficients $\vec{\beta}$ is calculated as
\begin{equation}
\vec{\beta}=U(T^T K U)^{-1} T^T \vec{y}.
\end{equation}
The columns $\vec{u}$ and $\vec{t}$ of matrices $U$ and $T$ are found iteratively from the KPLS algorithm [2]:
\begin{enumerate}
  \item Solve eigenproblem $(K\vec{y}\vec{y}^T)\vec{t}=\lambda\vec{t}$ for $\vec{t}$.
  \item Compute $\vec{u}=\vec{y}\vec{y}^T\vec{t}$.
  \item Deflate the kernel matrix $K \leftarrow (I-\vec{t}\vec{t}^T)K(I-\vec{t}\vec{t}^T)$.
\end{enumerate}
At each iteration, $\vec{u}$ and $\vec{t}$ are chosen so as to maximize the covariance between them [2]. The number of columns of $U$ and $T$ is equal to the number $\nu$ of latent variables of the model. Equivalently, the covariance between the projection of $\vec{y}$ onto $K$ is maximized.

A model is evaluated by comparing the prediction
\begin{equation}
\vec{z}=K\vec{\beta}
\end{equation}
against the known values $\vec{y}$, and can be assessed using the correlation coefficient:
\begin{equation}
r^2=1-\frac{\|\vec{y}-\vec{z}\|_2^2}{\|\vec{y}-E(\vec{y})\|_2^2}.
\end{equation}

\section{Implementation issues}

Data for each round was centered and scaled to zero median and unit absolute deviation.

For each round, PLS and KPLS algorithms were executed in Matlab using codes adapted from [11].

As the objective of the contest is to maximize the performance of the model for the prediction set, retained models from the calibration set must be robust. This is achieved by leave-one-out (LOO) cross-validation (CV). For a calibration set consisting of $\ell$ samples, this procedure involves using a single sample for validation and the remaining  $\ell-1$ samples for training. The training data is used to generate PLS/KPLS models and the validation data is used for model assessment. This is repeated $\ell$ times, such that each sample is used once for validation. Then, a correlation coefficient (Eq. 8) can be calculated from each sample's cross-validated prediction.

The framework of LOO CV permits the setting of the model hyperparameters. For PLS, the sole hyperparameters is the number $\nu$ of latent variables. For KPLS, both $\nu$ and the kernel parameter $\eta$ must be set. For each attempted combination of $\{\nu,\eta\}$, a calibration set LOO CV $r^2$ is obtained, and hyperparameter values are chosen so as to maximize the correlation coefficient.

But what values of $\{\nu,\eta\}$ are attempted? The number of latent variables is a positive integer, and we used the brute-force approach of trying all numbers between 1 and 20. As for $\eta$, it was optimized using MATLAB's simplex search provided by built-in routine \emph{fminsearch}.

\section{Contest modeling performances}

This section summarizes the modeling methods used in the three regression tasks.

For the first round, 584 RECON features were generated for each peptide. These features were used to supplement the provided COEPRA descriptors. The submitted model exploited Gaussian KPLS with a calibration set LOO CV $r^2$ of 0.7120. Contest results report an $r^2$ of 0.602 in the prediction set, a fourth-place finish.

For the second round, 147 RECON descriptors were generated to supplement the provided COEPRA descriptors for each sample. Gaussian KPLS resulted in a calibration set $r^2$ of 0.5799. This model gave a prediction $r^2$ of 0.735, a first-place result. Moreover, the $r^2$ was significantly higher than that of the second-place finisher at 0.612, by 20.1\%.

For the third round, 180 SIMIL descriptors were derived, corresponding to 20 descriptors per amino acid, which is the number of rows of a given column in the SIMIL similarity matrix. These descriptors were used for modeling in addition to the COEPRA ones. The exponential KPLS model was chosen this time, with $r^2=0.3737$ for LOO CV across the calibration set. Contest results report $r^2=0.201$ across the prediction set, a second-place finish.

\section{Further analysis}

Post-contest, it is possible to take a second look at the datasets, and perform more formal analyses on the COEPRA datasets. For example, freed from the tight deadlines within which submissions must be made, it is possible to optimize the kernel parameter $\eta$ to a higher level of accuracy, and try a greater number of combinations of the COEPRA, RECON and SIMIL descriptors. Despite the fact that the responses for the prediction set are now known, this analysis assumes that they are not for the purposes of model selection. Hence, model parameters $\{\nu,\eta\}$ are chosen based on LOO CV across the calibration set, as before, and a model is selected based on it's calibration LOO CV $r^2$ across the calibration set, and not from the prediction set.

Three questions emerged from contest results:
\begin{itemize}
  \item What was the improvement of using KPLS over that of PLS models?
  \item What was the value-added of using the 2D MOE and RECON autocorrelation descriptors (RAD)?
  \item What is the value-added of using the SIMIL scores?
\end{itemize}
To address question 1, models were generated using PLS, Gaussian KPLS and exponential KPLS. To address questions 2 and 3, consistent sets of 327 2D MOE and RAD features were generated. Then, models were generated using only the COEPRA descriptors, only the MOE/RAD descriptors, only the SIMIL descriptors, both the COEPRA and MOE/RAD, COEPRA and SIMIL, and all three sets of descriptors. Table \ref{table2} presents the results of these experiments.

\begin{landscape}

\begin{table}[tb!]													
\caption{Results of post-contest experiments. The model with the highest calibration set LOO CV coefficient of correlation is bolded. The model with the highest prediction set correlation coefficient is italicized.}\label{table2}													
\begin{center}													
\begin{tabular}{ccccccc}													
\hline													
method	&	\bf{PLS}	&		&	\bf{KPLS}	&		&	\bf{KPLS}	&		\\
kernel	&	\bf{linear}	&		&	\bf{Gaussian}	&		&	\bf{exponential}	&		\\
	&	\bf{calibration}	&	\bf{prediction}	&	\bf{calibration}	&	\bf{prediction}	&	\bf{calibration}	&	\bf{prediction}	\\
\hline													
\bf{round 1}	&		&		&		&		&		&		\\
COEPRA	&	0.625	&	0.455	&	0.726	&	0.678	&	0.721	&	0.691	\\
MOE/RAD	&	0.261	&	0.344	&	0.407	&	0.386	&	0.427	&	0.495	\\
SIMIL	&	0.512	&	0.352	&	0.575	&	0.549	&	0.583	&	0.618	\\
COEPRA+MOE/RAD	&	0.680	&	0.464	&	\bf{0.742}	&	\bf{0.661}	&	\it{0.724}	&	\it{0.694}	\\
COEPRA+SIMIL	&	0.620	&	0.459	&	0.735	&	0.664	&	0.721	&	0.693	\\
all	&	0.675	&	0.466	&	0.739	&	0.663	&	0.727	&	0.694	\\
\hline													
\bf{round 2}	&		&		&		&		&		&		\\
COEPRA	&	0.298	&	0.401	&	0.498	&	0.746	&	0.470	&	0.590	\\
MOE/RAD	&	0.095	&	0.144	&	0.323	&	0.546	&	0.301	&	0.441	\\
SIMIL	&	0.142	&	0.200	&	\bf{0.613}	&	\bf{0.427}	&	0.482	&	0.515	\\
COEPRA+MOE/RAD	&	0.293	&	0.403	&	\it{0.502}	&	\it{0.784}	&	0.464	&	0.591	\\
COEPRA+SIMIL	&	0.279	&	0.412	&	0.505	&	0.754	&	0.475	&	0.595	\\
all	&	0.275	&	0.414	&	0.509	&	0.782	&	0.469	&	0.596	\\
\hline													
\bf{round 3}	&		&		&		&		&		&		\\
COEPRA	&	0.302	&	0.153	&	0.354	&	0.200	&	0.373	&	0.219	\\
MOE/RAD	&	0.162	&	-0.135	&	0.104	&	0.035	&	0.177	&	0.200	\\
SIMIL	&	0.237	&	0.032	&	0.335	&	0.118	&	0.326	&	0.169	\\
COEPRA+MOE/RAD	&	0.303	&	0.178	&	0.354	&	0.212	&	\it{0.375}	&	\it{0.242}	\\
COEPRA+SIMIL	&	0.305	&	0.149	&	0.356	&	0.197	&	0.376	&	0.219	\\
all	&	0.305	&	0.173	&	0.356	&	0.208	&	\bf{0.377}	&	\bf{0.240}	\\
\hline													
\end{tabular}													
\end{center}													
\end{table}													

\end{landscape}						
												
For round 1, Table 2 shows that Gaussian KPLS with the combined COEPRA and MOE/RAD descriptors finds calibration and prediction correlation coefficients of 0.741 and 0.661, respectively. The latter comes very close to the contest's first-place result of 0.677. Note that a higher performance would have been achieved had the exponential kernel been chosen. However, the assumption is that only calibration set responses are known. Hence, the retained model must be based upon performance on the calibration set only. Also note that almost identical performances, within 0.010, are found if COEPRA+SIMIL or call descriptors are used.

For round 2, it is Gaussian KPLS, using only the SIMIL descriptors, that boasts the highest calibration set LOO CV $r^2$ with a value of 0.613. However, it seems that the model performance does not translate well to the prediction set, with an $r^2$ of 0.427. It is noticed that all other models have an improved prediction set correlation coefficient.

Ignoring that model for a moment, it is Gaussian KPLS with all descriptors that outperforms other models, with calibration and prediction $r^2$'s of 0.509 and 0.781, respectively. It appears that an increased number of MOE and RAD features in concert with a fine-tuning of the kernel parameter achieves a model that has $r^2$ of 0.046 (or 21.9\%) higher than this paper's first-place contest submission presented in the previous section. Note again that almost equal results are obtained for COEPRA+MOE/RAD, COEPRA+SIMIL and all descriptors.

For round 3, the exponential kernel performs better than the Gaussian kernel. Once again, both sets of inputs are used. Once again, results between COEPRA+MOE/RAD, COEPRA+SIMIL and all descriptors are quasi-identical. The best of the three has $r^2$ performances 0.375 and 0.242 across the calibration and prediction sets. This beats the contest winner by 0.006 or 2.5\%.

Looking back, it makes sense that the value-added of MOE/RAD and SIMIL are similar, since half of the weight of the SIMIL scores are based on MOE/RAD features.

\section{Conclusion}

\label{conclusion2}

Two conclusions stem from this paper.

First, in answer to question 1 relating to the possibility of improvement of KPLS over PLS, this paper finds that there is a very significant advantage in using nonlinear KPLS models over linear PLS ones.

Second, in answer to questions 2 and 3, while MOE/RAD descriptors or the SIMIL scores are insufficient to build performing models for the prediction of binding affinities, they contribute to improved modeling performance, in conjunction with the COEPRA descriptors. With further knowledge of the nature of the COEPRA descriptors, it may be possible to further specify the value-added contribution of the MOE/RAD and SIMIL features.

\section*{Acknowledgments}

This work was supported by NIH grant 1P20-HG003899-01. Charles Bergeron was supported by a doctoral fellowship from the \emph{Fonds quebecois de la recherche sur la nature et les technologies}. Margaret McLellan contributed a script used in sequence-to-structure conversion.

%\bibliography{../bibtex/a,../bibtex/b,../bibtex/c,../bibtex/d,../bibtex/efgh,../bibtex/ijkl,../bibtex/m,../bibtex/n,../bibtex/opqr,../bibtex/s,../bibtex/t,../bibtex/uvwxyz}
%\bibliographystyle{icml2010}

\section*{References}

[1] Wold, S., Ruhe, H., Wold, H. and Dunn III, W.J. (1984) The collinearity problem in linear regression. The partial least squares (PLS) approach to the generalized inverse. SIAM Journal of Scientific and Statistical Computations. 5:735-743.

[2] Rosipal, R. and Trejo, L.J. (2001) Kernel Partial Least Squares Regression in Reproducing Kernel Hilbert Space.  Journal of Machine Learning Research. 2:97-123.

[3] Doytchinova, I.A., Walhshe, Valerie, Borrow, Persephone and Flower, D.R. (2005) Towards the chemometric dissection of peptide HLA-A*0201 binding affinity: comparison of local and global QSAR models. Journal of Computer-Aided Molecular Design. 19:203-212.

[4] Hattotuwagama, C.K., Guan, Pingping, Doytchinova, I.A. and Flower, D.R. (2004) New horizons in mouse immunoinformatics: reliable in silico prediction of mouse class I histocompability major complex peptide binding affinity. Organic and Biomolecular Chemistry. 2:3274-3283.

[5] Doytchinova, I.A. and Flower, D.R. (2002) Physicochemical Explanation of Peptide Bonding to HLA-A*0201 Major Histocompatibility Complex: A Three-Dimensional Quantitative Structure-Activity Relationship Study. Proteins: Structure, Function and Genetics. 48:505-518.

[6] Breneman, C.M. and Rhem, M. (1997) A QSPR Analysis of HPLC Column Capacity
Factors for a set of High-Energy Materials Using Electronic Van der Waals
Surface Property Descriptors Computed by the Transferable Atom Equivalent Method.  J. Comput. Chem., 18(2), 182-197 .

[7] Breneman, C.M., Thompson, T.R., Rhem, M. and Dung, M. (1995) Electron Density
Modeling of Large Systems Using the Transferable Atom Equivalent Method, Computers \& Chemistry, 19(3), 161.

[8] Whitehead, C.E., Sukumar, N., Breneman, C.M. and Ryan, M.D. (2003) Transferable
Atom Equivalent Multi-Centered Multipole Expansion Method. J. Comp. Chem., 24: 512-529.

[9] Sukumar, N. and Breneman. C.M. (2007) QTAIM in Drug Discovery and Protein
Modeling in The Quantum Theory of Atoms in Molecules: From Solid State
to DNA and Drug Design. (C.F. Matta \& R.J. Boyd, Editors) Wiley-VCH.

[10] Katt, Bill. (2004) A Semi-Automated Approach To Molecular Discovery Through Virtual High Throughput Screening, Rensselaer Polytechnic Institute, Troy, New York.

[11] Shawe-Taylor, John and Cristianini, Nello. (2004) Kernel Methods for Pattern Analysis. Cambridge: Cambridge, UK.

\end{document}